\documentclass[pdflatex,sn-mathphys-num]{sn-jnl}

\usepackage{geometry}              
\usepackage{graphicx}%
\usepackage{multirow}%
\usepackage{amsmath,amssymb,amsfonts}%
\usepackage{amsthm}%
\usepackage{mathrsfs}%
\usepackage[title]{appendix}%
\usepackage{xcolor}%
\usepackage{textcomp}%
\usepackage{manyfoot}%
\usepackage{booktabs}%
\usepackage{algorithm}%
\usepackage{algorithmicx}%
\usepackage{algpseudocode}%
\usepackage{listings}%
\usepackage{hyperref}
\usepackage{caption}
\usepackage[none]{hyphenat}
\usepackage{cleveref} 

\theoremstyle{thmstyleone}%
\theoremstyle{thmstyletwo}%

\theoremstyle{thmstylethree}%
\begin{document}

\title[Article Title]{Differentiable Inverse Rendering Recovers Intrinsic Properties from Intravascular Optical Coherence Tomography}


\author*[1]{\fnm{Ajay} \sur{Manicka}}\email{ajay\_m@mit.edu}
\author[1]{\fnm{Karim} \sur{Kadry}}\email{kkadry@mit.edu}
\author[1,2]{\fnm{Sebastián} \sur{Ruiz-Lopera}}\email{srulo@mit.edu}
\author[1]{\fnm{Maryam} \sur{Alsharqi}}\email{maryam7@mit.edu}
\author[3]{\fnm{Sanjana} \sur{Arji}}\email{sarji@bwh.harvard.edu}
\author[1,4]{\fnm{Samuel} \sur{Ruip\'erez-Campillo}}\email{sruipere@mit.edu}
\author[1,2]{\fnm{Néstor} \sur{Uribe-Patarroyo}}\email{uribepatarroyo.nestor@mgh.harvard.edu}
\author[1,2]{\fnm{Brett E.} \sur{Bouma}}\email{bbouma@mit.edu}
\author[3]{\fnm{Farhad} \sur{Nezami}}\email{frikhtegarnezami@bwh.harvard.edu}
\author[1,3]{\fnm{Elazer R.} \sur{Edelman}}\email{ere@mit.edu}

\affil[1]{\orgdiv{Institute of Medical Engineering and Science}, \orgname{MIT}, \orgaddress{\street{77 Massachusetts Ave}, \city{Cambridge}, \state{MA}, \postcode{02139}, \country{USA}}}
\affil[2]{\orgdiv{Wellman Center for Photomedicine}, \orgname{Harvard Medical School and Massachusetts General Hospital}, \orgaddress{\street{399 Revolution Dr}, \city{Somerville}, \state{MA}, \postcode{02145}, \country{USA}}}
\affil[3]{\orgdiv{Brigham and Women's Hospital}, \orgname{Harvard Medical School}, \orgaddress{\street{75 Francis St}, \city{Boston}, \state{MA}, \postcode{02115}, \country{USA}}}
\affil[4]{\orgdiv{Department of Computer Science}, \orgname{ETH Zurich}, \orgaddress{\street{R\"amistrasse 101}, \city{Zurich}, \postcode{8092}, \country{Switzerland}}}


\abstract{\unboldmath Medical images encode intrinsic tissue properties that indicate disease progression, yet these properties remain entangled with the system-dependent physics of image formation. Differentiable inverse rendering asks whether they are recoverable by inverting the imaging system's forward physics rather than through post-hoc correction. We address this in optical coherence tomography (OCT), focusing on the optical attenuation coefficient, a per-pixel marker of pathology and cellular composition. Conventional recovery relies on closed-form inversions limited to simplified light-tissue models, or learned methods coupled to specific imaging configurations. We instead recover per-pixel attenuation from a single OCT image by gradient descent through a differentiable physics-based forward model. We verified the method against analytical solutions, validated it on five in vitro phantoms, and applied it without modification to 29{,}269 clinical intravascular OCT images from 128 patients. The inversion needs no training data or segmentations and is portable across platforms and retrospective archives. Replacing the forward model or recovering another optical parameter leaves the optimization procedure unchanged. We demonstrate per-pixel recovery is tractable because gradient information scales to this high-dimensional problem, where gradient-free methods become intractable. Our work supports quantitative tissue characterization from OCT images and provides a framework for extension across imaging modalities and organ systems.}




\maketitle
\newpage
\section*{Introduction}\label{sec:Introduction}

The interrogation of biological matter through diverse forms of energy constitutes the foundation of modern medical diagnostic practice. Whether utilizing magnetic gradients in magnetic resonance imaging \cite{berger2002does}, high-frequency sound waves in ultrasound \cite{harvey2002advances}, ionizing radiation in computed tomography \cite{buzug2011computed}, or backscattered light in optical coherence tomography (OCT) \cite{drexler2008optical}, modern imaging modalities share a common operative principle. They apply energy to tissue and infer anatomy from how the energy is differentially reflected, scattered, or absorbed by the underlying material \cite{shung2012principles}. Historically, the advancement of these technologies has focused on hardware innovations to improve spatial resolution and contrast, providing clinicians with high-fidelity anatomical visualizations of organ systems and diseased tissue \cite{webb2022introduction,oppelt2006imaging}. However, while these anatomical maps are essential, they are not direct measurements of tissue composition: each image is a projection of the underlying intrinsic properties through the physics of the imaging system, entangling composition with modality-specific acquisition artifacts. In clinical applications, intrinsic properties such as stiffness, conductivity, and pH dictate disease progression and physiological stability \cite{levental2009matrix,webb2011dysregulated,gabriel1996dielectric}, yet remain inaccessible from the projected image alone. This paper inverts that projection. Although the imaging chain is a complex, modality-specific forward operator whose inverse is ill-posed, we demonstrate intrinsic tissue composition can be recovered directly from clinical imaging through differentiable inverse rendering, with OCT as our case study.

The technical foundation for this inversion comes from parallel development in computer vision. Physics-based forward models have been refined in medical imaging \cite{webb2022introduction}, yet they are almost exclusively used in alternate directions. They simulate, correct, or render images rather than recover tissue properties from them. The computer vision community is closing an analogous gap through differentiable rendering \cite{li2018differentiable, pharr2023physically}, which treats the image-formation process as a differentiable function and uses gradient-based optimization to recover scene parameters from observed images \cite{mildenhall2021nerf}. The principle that gradient descent can invert any differentiable forward model to recover its input extends naturally beyond natural images and has begun to enter medical imaging in modality-specific applications such as x-ray and ultrasound \cite{gopalakrishnan2024intraoperative, bertramo2025diffus, finkelstein2025multi, simson2023differentiable}.

In biomedical optical imaging, the optical attenuation coefficient (OAC) is a particularly informative quantitative tissue property, characterizing how rapidly the signal is attenuated with depth. Attenuation serves as a robust discriminator of pathology across both acoustic \cite{kuc2007clinical,berger1990global,sagar1990quantitative,sugata1992application} and optical \cite{xu2008characterization,mclaughlin2010parametric,barwari2011advanced,barwari2012differentiation,tomlins2010scattering,zhang2012quantitative,van2012effect} regimes, and is sensitive to fundamental cellular mechanisms, including necrosis and apoptosis \cite{van2010apoptosis}, making it a functional proxy for tissue viability beyond the shape and size of anatomical features captured by standard imaging. While IV-OCT provides high anatomical resolution \cite{huang1991optical,bouma2017intravascular}, the image entangles the OAC with the effects of complex wave propagation and system-specific acquisition parameters, making it difficult to extract intrinsic composition from image intensity \cite{stone2011prospective,virmani2000lessons}.

Other biomedical optical modalities recover different optical properties. Photoacoustic and diffuse optical methods derive contrast from optical \emph{absorption}, where quantitative recovery is itself an ill-posed problem because the internal fluence is unknown \cite{wang2012photoacoustic, cox2012quantitative}, and intravascular photoacoustic catheters remain preclinical \cite{iskander2019photoacoustic}. OCT attenuation is instead dominated by scattering, the mechanism carrying lipid, calcium, and cellular density information in plaque \cite{yabushita2002characterization, van2010atherosclerotic}, and is well established as a quantitative marker across many tissue types \cite{scolaro2012parametric, cauberg2010quantitative, gong2014assessment, qin2021research, almasian2015validation}. Because IV-OCT is already standard of care during percutaneous coronary intervention \cite{tearney2012consensus, yonetsu2024cardiac}, inverting its image formation extracts an additional quantitative channel from images that are already acquired. OCT is a case study here as the framework requires only that the forward operator be differentiable with respect to the property of interest.

Existing approaches to OCT attenuation recovery fall into three categories \cite{chang2019review}. Curve-fitting (CF) methods extract a single attenuation coefficient by fitting signal decay over a manually segmented region \cite{van2005localized}, which is robust for uniform tissues but unable to capture intra-tissue heterogeneity. Depth-resolved (DR) methods compute per-pixel attenuation from the ratio of local intensity to remaining energy \cite{vermeer2013depth, liu2017tissue}, but the closed-form formulation constrains them to single-scattering assumptions and makes the resulting maps sensitive to noise sources like speckle and local signal artifacts. Recent work in this area addresses the noise sensitivity issue using despeckling to improve local region homogeneity~\cite{cannon2021layer}. Learning-based methods can incorporate richer forward physics through synthetic training data \cite{yu2026endtoend}, but the learned amortized inverse is fixed after training and coupled to the simulated distribution. The stationary iteration method estimates depth-dependent attenuation and backscattering without explicit layer segmentation~\cite{wang2023depth}, and optimization also estimates these parameters via an implicit neural representation of optical properties through an OCT imaging model using multi-angle measurements~\cite{li2025computational}. Our method instead optimizes any differentiable forward model by gradient descent from a single B-scan.

Our work targets a central question in differentiable inverse rendering of whether intrinsic tissue properties can be inferred directly from medical images by framing property recovery as the inverse of the imaging system's forward physics. To examine this in OCT, we built an automated and generalizable physics-based inverse rendering framework that extracts intrinsic tissue properties directly from standard clinical OCT images without the need for segmentations. The framework addresses all three limitations in a single formulation by delivering per-pixel attenuation while operating on an explicit forward model that can incorporate arbitrary physics and optimizing each image independently. Because the forward model is modular, incorporating additional scattering physics, such as the Extended Huygens-Fresnel model \cite{thrane2000analysis}, or recovering further optical parameters requires only modifying the forward model, not the inversion procedure.

By inverting the path-integrated Beer-Lambert (B-L) model \cite{oshina2021beer}, we recovered depth-resolved, pixel-level OAC maps. We verified our method on five optically homogeneous physical phantoms against the conventional CF method and applied it to 29{,}269 clinical coronary IV-OCT frames~\cite{ali2017optical,ali2019safety,kereiakes2023intravascular,saito2022intravascular} (over a 128-patient intravascular lithotripsy cohort) to demonstrate spatially resolved attenuation recovery and faithful image rendering across varying vessel morphologies. IV-OCT routinely guides percutaneous coronary intervention. Images are acquired with a catheter-based scanner that illuminates the arterial wall with a near-infrared laser and reconstructs depth-resolved cross-sections from the backscattered light. Current clinical practice relies on viewing anatomical features such as plaque shape and size, while treatment decisions depend on tissue composition. These results support the potential of per-pixel OAC maps to enable quantitative plaque characterization and guide interventional decision-making~\cite{van2010apoptosis,fujino2018new}.

The differentiable inverse rendering approach generalizes to any imaging modality whose forward physics can be expressed as a differentiable operator. Changing the forward physics can change the properties that are recovered. We can also control the optimization loop, loss function, and regularization framework to guide optimization. Our work demonstrates this paradigm in OCT, joining differentiable forward models already demonstrated for other modalities across physics-based medical imaging \cite{chen2025differentiable, barbastathis2019use}. We make the following contributions:
\begin{itemize}
    \item A gradient-based differentiable inverse rendering framework for per-pixel optical attenuation recovery from OCT images without training data or prior segmentation. 
    \item Quantitative validation against curve-fitting references across 5,120 B-scan inversions on 5 physical liquid intralipid phantoms each imaged at two focal positions, with systematic characterization of optimizer reproducibility, noise robustness, and parameter sensitivity in agar phantoms.
    \item Retrospective application to 128 clinical IV-OCT coronary pullbacks of atherosclerotic tissue demonstrating spatially resolved attenuation and backscattering recovery. 
\end{itemize}
Differentiable inverse rendering posits that intrinsic property recovery from images can be cast as a tractable inverse problem driven by gradient information. Our work contributes evidence for this paradigm in biomedical imaging, with OCT as the use case. We introduce a differentiable inverse rendering framework for OCT with validation across phantoms and demonstration in clinical acquisitions. Gradient information is vital because it captures the intrinsic relationships between tissue regions and guides the optimizer toward a physically consistent solution. The inversion is portable across different imaging systems and applicable to retrospective clinical archives, addressing the domain-shift barrier in clinical AI deployment. Gradient descent through an explicit forward model decouples the complexity of the physics from the availability of an inverse, whereas closed-form estimators require physics simple enough to invert analytically and learned estimators are bound to their training distribution, making the forward model a replaceable component and the approach transferable beyond OCT.


\section*{Results}\label{sec:Results}

Our method takes a single OCT image as input along with prespecified scanner and optimization parameters. It returns a per-pixel map of the optical attenuation coefficient $\mu$. An OCT image is composed of depth profiles called A-lines; a two-dimensional cross-section (B-scan) is formed by laterally concatenating adjacent A-lines. The framework operates by passing a candidate attenuation map through a differentiable forward model that simulates a simplified OCT image by combining Beer-Lambert attenuation with confocal amplitude modulation, then iteratively updating $\mu$ by gradient descent to minimize the discrepancy between the simulated and measured B-scan. We first verify the forward and inverse components on synthetic data before applying the framework to physical phantoms and clinical intravascular OCT images.

\subsection*{Forward and inverse models achieve exact recovery on computational phantoms}\label{sec:fwdInvVerification}

We compared four optimizers while increasing the number of axially coupled attenuation parameters within one synthetic A-line from 25 to 5,000 across eight problem sizes (\Cref{fig:optimizer_baseline}). All methods shared the forward model, bounds, and eight paired initializations, with solver-specific optimization budgets. All achieved below 0.1\% mean MAPE at the smallest size. While Adam and L-BFGS-B maintained this accuracy throughout, Random Search and CMA-ES became worse reaching MAPE values of 30.32\% and 100.91\% respectively. Supplementary experiments examined lateral scaling on synthetic data, measured-data recovery, and A-line coupling (Supplementary Figure~1). At 475 parameters on the measured A-line with axial TV, the gradient-based methods achieved lower intensity MSE and closer CF agreement. With fixed gain and no lateral TV, each A-line can be optimized independently using the mean squared error between predicted and measured intensities and the axial TV regularizer that couples neighboring pixels in a single A-line together by penalizing differences in attenuation. Lateral TV couples neighboring A-lines together within a B-scan. Estimating one gain from the entire B-scan also couples the A-lines because each A-line contributes to the shared gain. We therefore optimize all A-lines in a B-scan together in the agar-phantom and clinical inversions.

The differentiable forward model was first verified against the analytical Beer-Lambert solution. Five hundred synthetic A-lines, each containing 2 to 5 homogeneous layers with randomized attenuation coefficients spanning $[0.1, 15.0]$~mm$^{-1}$ over a depth range of 0.4--0.8~mm (100 depth pixels at $\Delta z = 4$~\textmu m; \Cref{fig:verification}g), were passed through the differentiable forward model to produce the synthetic OCT B-scan (\Cref{fig:verification}h) and through the analytical Beer-Lambert solution to produce the target OCT B-scan. Representative A-line profiles for 2-, 3-, and 4-layer configurations confirm numerical agreement (\Cref{fig:verification}a--f). Across all 50,000 pixels, the residuals were five orders of magnitude below the lowest signal ($R^2 = 1.0000$, MAE $= 5.0 \times 10^{-6}$~dB, max error $= 6.1 \times 10^{-5}$~dB; \Cref{fig:verification}i), indicating virtually exact agreement.

The inverse optimization was then evaluated on 500 synthetic B-scans generated by the forward model from randomized ground-truth attenuation maps, testing whether the optimizer can recover known attenuation maps from its own rendered image. We show one example B-scan optimization result (\Cref{fig:verification}j--o), concordance scatter plots at four optimization snapshots across all 500 phantoms (\Cref{fig:verification}p--s), and the full convergence trajectory of MAPE and $R^2$ over 30{,}000 iterations (\Cref{fig:verification}t). Each attenuation map was initialized from scrambled conditions ($R^2 < 0.0001$, MAPE $= 77.0\%$; \Cref{fig:verification}j,p), and the optimizer converged monotonically to near-perfect recovery (\Cref{fig:verification}t). Across all 500 phantoms, the recovered tissue $\mu$ achieved final MAPE~$= 0.15\%$, concordance $R^2 = 0.9997$, MAE $= 0.007 \pm 0.001$~mm$^{-1}$, and regression slope $= 0.998$. These results confirm the numerical correctness of both the forward operator and the optimization procedure, establishing a baseline for evaluation on physical phantoms. Recovery under forward-model mismatch is evaluated in Supplementary Note~7.

\subsection*{Recovered attenuation tracks scatterer concentration across physical phantoms}\label{sec:phantomValidation}

Physical phantoms provide a controlled setting in which the imaging system parameters are fully specified and reference attenuation values are available from conventional curve-fitting analysis, enabling direct quantitative comparison with the optimization-based recovery. All 512 B-scans per acquisition were processed independently using the single-scattering forward model, totaling 5{,}120 B-scan inversions across five liquid intralipid phantoms (0.5--3.0\%) each imaged at two focal positions (H and L; \Cref{fig:phantom_boxplot}a--c). The optimization workflow on a representative 5.0\% intralipid-in-agar B-scan is shown in Supplementary Figure~2d--i, with the recovered $\mu$ volume for the 3.0\% H liquid acquisition (\Cref{fig:phantom_boxplot}d,e) demonstrating spatially coherent attenuation within the tissue band.

Recovered attenuation coefficients, evaluated within matched CF windows, increased monotonically with intralipid concentration across all 10 acquisitions (\Cref{fig:phantom_boxplot}f). Frame-to-frame variability was low (standard deviation $0.025$--$0.053$~mm$^{-1}$ across 512 B-scans per volume). Each concentration was imaged at two focal positions; H and L yielded slopes of $0.663$ and $0.629$~mm$^{-1}$\% with $R^2 = 0.990$ and $R^2 = 0.993$, comparable to the CF reference (slopes of $0.623$ and $0.625$~mm$^{-1}$\%). The framework recovers a full per-pixel OAC map across the tissue volume (\Cref{fig:phantom_boxplot}e), whereas CF collapses each spatially averaged measurement location to a single slope value. Absolute values were $0.083$--$0.176$~mm$^{-1}$ below CF, but the linear relationship and relative ordering are preserved. Example A-lines (\Cref{fig:phantom_boxplot}g) show lower intensity RMSE for DiffOCT than for CF ($0.29$ and $0.52$ versus $0.95$ and $1.76$~dB), although attenuation agreement remains conditional on the shared CF-derived gain.

\subsection*{Attenuation recovery is reproducible across initializations and robust to noise}\label{sec:identifiabilityRobustness}

The identifiability study quantified the reproducibility of the recovered attenuation under two distinct perturbations of the initialization: the random draw of the initial $\mu$ map from a fixed distribution (seed sweep), and the width and location of the initialization distribution itself (range sweep) (\Cref{fig:ident_robust}a--d). In the seed sweep, each acquisition was optimized 20 times with independent initializations drawn from a fixed uniform distribution (\Cref{fig:ident_robust}a). Intra-acquisition coefficients of variation (CoV) ranged from $0.18\%$ to $0.32\%$ across the 10 acquisitions (mean $= 0.27 \pm 0.04\%$; \Cref{fig:ident_robust}c), indicating that, given a fixed initialization distribution, the optimizer yields reproducible tissue-mean attenuation.

In the range sweep, 10 distinct ranges for the uniform distribution were tested with 5 seeds per condition (\Cref{fig:ident_robust}b), spanning narrow ($[3, 7]$~mm$^{-1}$), wide ($[0, 10]$, $[0, 20]$~mm$^{-1}$), fixed-value ($[1, 1]$, $[5, 5]$, $[10, 10]$, $[15, 15]$~mm$^{-1}$), biased ($[7, 15]$~mm$^{-1}$), and informed (CF reference) conditions. Seven of the 10 conditions yielded consistent recovery across all 10 acquisitions: after averaging the five seeds per condition, every condition mean was within $\pm25\%$ of the corresponding acquisition's mean across these seven conditions. The corresponding per-acquisition initialization range CoV was $2.1\%$--$13.1\%$ (\Cref{fig:ident_robust}c,d). The three conditions $[0, 3]$, $[1, 1]$, and $[15, 15]$~mm$^{-1}$ failed this criterion and yielded different recovered values, consistent with the non-convexity expected from the cumulative-sum forward operator. This criterion describes numerical consistency rather than accuracy, since pixel-wise ground truth is unavailable for measured data. Initialization-induced uncertainty in the clinical maps is quantified in Supplementary Note~8.

The noise robustness study quantified how additive Gaussian noise degrades recovery across all 10 acquisitions over a sweep from SNR $= -3$~dB up to each phantom's baseline acquisition SNR in 1~dB steps (\Cref{fig:ident_robust}e--g). Following the noise model in OCT~\cite{uribe2020noise}, zero-mean circular Gaussian noise was added to each complex-valued B-scan at a controlled SNR computed using the rigorous definition that accounts for the noise floor bias in the numerator. For every phantom, three recovery metrics tracked SNR consistently. Recovered tissue $\mu$ deviated only marginally from the noiseless baseline over the first 4--5~dB of added noise and then diverged sharply as SNR fell further (\Cref{fig:ident_robust}g). Baseline concordance $R^2$ (recovered $\mu$ versus its noiseless recovery) increased monotonically with SNR across every phantom. The spatial standard deviation of recovered tissue $\mu$ decreased monotonically with SNR, reflecting smoother and more spatially consistent recoveries at high SNR.

\subsection*{Recovered attenuation maps generalize to clinical intravascular OCT}\label{sec:clinicalApplication}


To demonstrate the method transfers from phantoms to clinical acquisition, the framework was applied to intravascular OCT with the same optimization algorithm and reconstruction loss, without TV regularization. Per-frame optimization was run independently on 29{,}269 cross-sections spanning 128 separate patient pullbacks drawn from four clinical trial cohorts~\cite{ali2017optical,ali2019safety,kereiakes2023intravascular,saito2022intravascular}; each Cartesian frame was converted to a polar representation for optimization. Each frame is solved independently, as the method carries no state or learned parameters between frames. \Cref{fig:clinical} shows 5 example frames from a single representative pullback. Because scanner-specific calibration metadata was unavailable for this acquisition, we adopted measured confocal parameters reported for the C7-XR / Dragonfly catheter family ($z_R = 0.59$~mm, $z_f = 1.06$~mm)~\cite{gargesha2015parameter}, together with an axial pixel size of $0.01932$~mm/pixel estimated from the ILUMIEN OPTIS A-scan range. Combined with the undocumented post-processing transformations applied to the log-compressed clinical images, the recovered attenuation values capture relative tissue contrast rather than absolute quantities.

The per-pixel attenuation coefficient $\mu$ was first optimized with the backscattering fraction $\alpha$ held fixed at unity (Stage~1) assuming a 15 dB noise floor. Across every frame of every patient, the recovered OAC maps are spatially resolved and track the tissue architecture visible in the raw images, resolving the vessel wall, lumen boundary, and calcified plaque deposits, confirming that the framework generalizes from controlled phantom to the complex radial scan of catheter-based IV-OCT. Using the recovered $\mu$ maps, the forward model synthesizes rendered IV-OCT images that reproduce the overall geometry of each cross-section. Aggregated over all 29{,}269 frames, rendered and target B-scans agree closely on tissue coverage (Dice $= 0.957 \pm 0.007$), and pixel-level rendering quality reaches a mean PSNR of $14.83 \pm 0.99$~dB (\Cref{eq:psnr}).

The Stage~1 residual maps (right column of \Cref{fig:clinical}) reveal structured, non-random differences concentrated along the inner wall. These arise because $\mu$ plays a dual role in the Beer-Lambert forward model as in \Cref{eq:dB_fwd_discrete}: it controls per-pixel brightness and cumulative attenuation, so the optimizer must satisfy both simultaneously with a single variable. Introducing a per-pixel backscattering parameter $\alpha$ (Stage~2) decouples these roles: $\mu$ is frozen and $\alpha \in [10^{-10}, 10^{-2}]$ is optimized as an independent brightness term. This reduces residuals across the cohort and raises the mean PSNR to $22.63 \pm 0.74$~dB (Dice $= 0.954 \pm 0.009$; \Cref{fig:clinical}, bottom row) without altering the recovered attenuation structure. A detailed derivation of the brightness-attenuation coupling and the two-stage procedure is provided in Supplementary Note~6.
The recovered maps carry information that the standalone IV-OCT image does not. Pixel brightness is not a tissue property. It confounds local backscattering with the cumulative attenuation of every voxel already traversed, the confocal gate, and an arbitrary system gain such that identical tissue appears at different intensities depending on depth and position relative to the catheter. Inverting the forward model removes these confounders so equal $\mu$ denotes equal optical property anywhere in the frame.

No pixel-level ground-truth attenuation exists for in vivo coronary tissue, so this experiment demonstrates that the method transfers unmodified to clinical acquisition geometry rather than validating recovered attenuation values. PSNR and Dice quantify forward-model fidelity, not the accuracy of $\mu$, which is demonstrated with the computational and physical phantoms.

\section*{Discussion}\label{sec:Discussion}
We investigated whether intrinsic tissue properties can be recovered directly from medical images by recasting property extraction as an inverse problem that explicitly accounts for the system-dependent effects of image formation rather than a post-processing task. We tested this hypothesis in OCT by developing a differentiable inverse rendering framework that recovers the optical attenuation coefficient (OAC) per pixel from a single OCT image, without training data or prior segmentation. By embedding the Beer-Lambert model as a differentiable forward operator and optimizing $\mu$ directly via gradient descent, the framework produces per-pixel parameters with direct physical meaning and an optimization procedure decoupled from the imaging configuration. 

The optimizer comparison supports gradient-based attenuation recovery as the number of axially coupled parameters increases (\Cref{fig:optimizer_baseline}). All four methods achieve low error at small problem sizes, but the gradient-free algorithms lose accuracy as dimensionality grows. Random Search runs only a forward pass, so its computational cost stays flat, but its error grows steadily with problem size. CMA-ES fails on both fronts. Its error also grows with dimensionality, and each evaluation updates a covariance matrix whose size grows as the square of the number of parameters, making it computationally prohibitive at large sizes. Adam adds only a backward pass to compute gradients, while L-BFGS-B additionally maintains a short history of recent gradient and position updates to build its quasi-Newton search direction. Only the gradient-based methods remain viable across the full scaling range, underscoring the necessity of gradient information for high-dimensional inverse problems in imaging.

We then verified numerical correctness on synthetic phantoms (\Cref{fig:verification}), and we examined 5,120 inversions on liquid intralipid phantoms and found that recovered OAC tracks scatterer concentration linearly ($R^2 = 0.990$--$0.993$) with slopes of $0.629$--$0.663$~mm$^{-1}$\% under CF-derived gain calibration (\Cref{fig:phantom_boxplot}), and that agar-phantom recovery is reproducible across initializations (CoV = $0.18\%$--$0.32\%$; \Cref{fig:ident_robust}). The noise robustness study confirmed that recovery degrades predictably with decreasing SNR: across every phantom, baseline concordance and per-pixel spatial consistency tracked SNR monotonically, while recovered $\mu$ deviated only marginally from its noiseless value over the first few dB of added noise and diverged sharply only once noise approached the signal amplitude. The physics-based forward model therefore acts as a stable regularizer across realistic acquisition SNRs and loses this regularization only when the signal itself is lost. The OCT inverse problem is fundamentally ill-posed, meaning that many distinct attenuation maps can produce nearly identical observed images. Our reproducible recovery across initializations indicates that the physics-based forward model yields consistent tissue-mean attenuation within the initialization ranges examined, and agreement with CF in liquid intralipid phantoms (\Cref{fig:phantom_boxplot}) supports the accuracy of tissue-mean recovery.

Applying the same algorithm to clinical coronary IV-OCT without modification recovered spatially resolved OAC maps (\Cref{fig:clinical}), demonstrating that physics-based recovery transfers across acquisition geometries in a data-agnostic way that learned methods cannot achieve without retraining. The recovered maps produce synthesized images that match the acquired B-scans with a mean PSNR of $14.83 \pm 0.99$~dB, a modest value explained by the dual role that the OAC in the Beer-Lambert forward model plays in both per-pixel brightness and cumulative attenuation along depth. The optimizer must find a single $\mu$ map that simultaneously satisfies both constraints. Introducing a per-pixel backscattering parameter $\alpha$ decouples these roles and raises the mean PSNR to $22.63 \pm 0.74$~dB (Supplementary Note~6) without altering the recovered attenuation structure. This sequential procedure, constrained by literature-informed bounds on $\mu$ and $\alpha$~\cite{chang2019review,cannon2021layer}, reduces but does not remove the ambiguity between attenuation and backscattering in a single intensity image. The modest $\mu$-only PSNR therefore reflects the under-specification of the single-parameter forward model rather than a failure of the inversion procedure. Despite lack of absolute calibration, the recovered maps preserve spatial contrast consistent with anatomical features, which is the primary signal used in clinical decision-making.

The ability to extract a well-established tissue biomarker \cite{van2005localized, van2010atherosclerotic} directly from routine IV-OCT imaging has direct implications for interventional cardiology, where treatment decisions hinge on tissue composition that is often inferred indirectly from anatomical features observed in imaging. Attenuation-based plaque assessment has already reached commercial deployment in the form of the OCT-derived index of plaque attenuation, which discriminates high-risk plaque features in vivo against expert consensus OCT classification \cite{cai2025vivo}, and this supports that the underlying signal carries clinical meaning. Such systems are, however, closed and vendor-bound pipelines that reduce a region to a single aggregate index, whereas the framework presented here recovers attenuation per pixel through an explicit forward model that can be substituted to recover other optical properties without altering the inversion. Per-pixel OAC maps could serve as a quantitative substrate for decisions that depend on spatial contrast, including vulnerable-plaque identification, calcified-region delineation for lithotripsy or atherectomy planning, and objective monitoring of plaque evolution across serial acquisitions of the same patient. Because the inversion depends on forward physics rather than a learned prior tied to specific scanners and samples, such analyses can be applied retrospectively to existing IV-OCT archives. These applications are most directly enabled once absolute calibration is available; the results here demonstrate the spatially resolved intrinsic properties and the relative contrast that would underpin them.

We observed a systematic underestimation of recovered attenuation relative to curve-fitting references, but the difference was small at every concentration ($0.083$--$0.176$~mm$^{-1}$), and concentration slopes were close to CF under shared gain calibration (\Cref{fig:phantom_boxplot}f). Both methods share the homogeneous Beer-Lambert limit, but CF fits a single attenuation to each depth window, acting as a first-order approximation of the signal decay, while DiffOCT fits an attenuation value at every depth pixel under TV regularization. The DiffOCT prediction therefore tracks the measured signal closely, whereas the CF line behaves as a regression line through it, with lower intensity RMSE for DiffOCT in both example A-lines (\Cref{fig:phantom_boxplot}g). The small offset is consistent with this difference in model order. In these homogeneous samples, part of the local variation that DiffOCT follows reflects residual speckle and noise, so lower RMSE indicates closer signal reproduction rather than more accurate attenuation by itself. DiffOCT thus reproduces the CF attenuation to within $0.2$~mm$^{-1}$ while additionally resolving attenuation with depth, which a single slope cannot represent in heterogeneous tissue. Because the loss mask is fixed from measured intensities, the offset does not arise from predictions falling below the noise floor and being excluded from the loss. Such systematic offsets are pervasive in the existing OAC retrieval literature. Reported values for calcified plaque alone span an order of magnitude depending on the estimation method, from $11.1 \pm 4.9$~mm$^{-1}$ by curve fitting on averaged A-scans \cite{van2005localized} to $0.87 \pm 0.24$~mm$^{-1}$ by depth-resolved analysis \cite{liu2017tissue}, yet the relative ordering of tissue types is preserved within each study \cite{chang2019review}. Tissue differentiation depends on the relative ordering of the OAC across tissues rather than on absolute magnitudes, and our recovered OAC preserves this ordering through a linear relationship to intralipid concentration across the phantom series. This has important implications for clinical settings. When scanner-specific calibration metadata is unavailable, recovered OAC values are no longer absolute but retain the within-slice relative contrast needed for tissue discrimination. Methods to convert relative OAC into absolute cross-patient values, such as reference-tissue anchoring or self-calibrating forward models, will be vital for extending the framework to cases requiring population-level comparisons such as drug-trial endpoints.

Solving this inversion at clinically relevant scales is tractable only because of the structural advantage of gradient information. Each backward pass delivers one derivative per parameter, while gradient-free methods update based on a scalar loss whose information content does not scale with problem size (\Cref{fig:optimizer_baseline}), which explains the convergence failure in gradient-free approaches with increasing problem size. Beyond this computational tractability, physics-based differentiable inverse rendering carries two deployment advantages over analytical and data-driven methods that are particularly relevant for digital health. First, the framework is fully modular. The forward model, optimizer, and regularization terms can each be modified independently without restructuring the inversion procedure. This flexibility is unavailable to closed-form analytical approaches, which require the forward physics to admit an analytical inverse, and to learning-based approaches, which are coupled to the physics of their training distribution. Second, because the inversion is agnostic to acquisition configuration, it naturally sidesteps the scanner-to-scanner domain shift that is the dominant barrier to clinical deployment, allowing a single implementation to serve heterogeneous clinical environments regardless of scanner vendor, acquisition protocol, or post-processing pipeline. As physics-based differentiable models become increasingly common across medical imaging, including ultrasound \cite{simson2023differentiable, bertramo2025diffus}, X-ray \cite{gopalakrishnan2024intraoperative}, and MRI \cite{finkelstein2025multi}, their shared portability and interpretability provide a foundation for quantitative imaging tools that can scale across the infrastructure of digital health.

However, while our framework can recover per-pixel OAC from clinical OCT images without retraining, it has several limitations. First, full optimization is computationally costly (approximately 35--50~s per B-scan on a modern GPU; Supplementary Note~4), which is tractable for offline research but incompatible with real-time catheter-lab use. Second, the clinical demonstration was conducted without access to scanner-specific system parameters or post-processing metadata, so the recovered values should be interpreted as evidence of geometric transfer and relative attenuation extraction rather than population-level clinical validation of absolute OAC values. This can be overcome in future developments by having access to the raw data and optical specifications of the scanner. Related to this, no reference standard for per-pixel attenuation exists in vivo, since histology is destructive and cannot be registered to individual OCT pixels. Third, our forward model is a simplified intensity-domain formulation that omits phase, coherent interference, and the speckle realization that arises from them. We adopt this simplification deliberately because phase-resolved simulation requires scanner-specific spectral parameters that are unavailable for retrospective clinical data. Speckle is nonetheless pervasive in IV-OCT and, because $\mu$ is fitted at every pixel, residual speckle can appear as local attenuation fluctuations (\Cref{fig:phantom_boxplot}g). In the phantom experiments, we mitigate this using Gaussian spatial filtering applied to input B-scans, which other despeckling methods can replace, and an anisotropic total-variation penalty (Eq.~\eqref{eq:total_loss}) that encourages spatial homogeneity within tissue regions. Fourth, tissue-mean recovery is reproducible across seeds (CoV $= 0.18\%$--$0.32\%$) and the seven selected initialization conditions (CoV $= 2.1\%$--$13.1\%$; \Cref{fig:ident_robust}c--d), but more aggressive perturbations of the initialization range produce different recovered values. Initialization-induced uncertainty in the clinical tissue means and individual pixels is quantified in Supplementary Note~8. Lastly, the single-scattering Beer-Lambert forward model may become insufficient at higher optical depths and in cases where multiple scattering dominates, although preliminary comparisons with the Extended Huygens-Fresnel model ($g = 0.7$, a value previously used for tissue in OCT simulations~\cite{hartinger2014monte}) yielded nearly identical recovered attenuation at our phantom concentrations (mean difference $<1\%$; Supplementary Note~5).

This work highlights several directions for extension. Posterior sampling and Bayesian inverse-problem formulations can attach per-pixel uncertainty bounds to recovered OAC and characterize the local-minima behavior we observed under aggressive initialization perturbation. More thorough treatment of multiple-scattering forward models can extend quantitative recovery into deeper tissue regimes, building on the Extended Huygens-Fresnel implementation reported in Supplementary Note~5. Joint estimation of system parameters such as $z_f$ and $z_R$ with $\mu$ is a natural extension of the framework and should be feasible with constraints that make these parameters identifiable.

This framework permits quantitative tissue characterization from routine clinical intravascular imaging using physics-based differentiable inverse rendering. By recovering physically interpretable attenuation maps from a single IV-OCT image without training data or scanner calibration, it can support interventional cardiologists in plaque characterization and treatment planning, and enable retrospective analysis of existing IV-OCT archives.


\section*{Methods}\label{sec:Methods}

\subsection*{Framework Overview}\label{sec:framework_overview}

The framework recovers per-pixel optical properties from a measured OCT B-scan by iteratively optimizing a differentiable forward model to reproduce the observed image (Fig.~\ref{fig:framework}). Given a log-scaled target B-scan $I_\text{target}(z, x)$, where $z$ denotes depth and $x$ the lateral A-line index, the method initializes a spatial map of tissue attenuation coefficients $\mu(z, x)$ and passes it through a physics-based forward model $\mathcal{F}$ that simulates OCT image formation, producing a predicted B-scan $\hat{I}(z, x) = \mathcal{F}(\mu; \boldsymbol{\theta})$, where $\boldsymbol{\theta}$ denotes the fixed system parameters (focal geometry, pixel size, noise floor, and system gain). A masked mean squared error (MSE) loss, with total variation (TV) regularization on $\mu$ for spatial smoothness, is computed between $\hat{I}$ and $I_\text{target}$ over tissue pixels identified by a noise floor mask. Because the forward model is implemented as a differentiable computational graph in PyTorch \cite{paszke2017automatic}, the gradient $\partial \mathcal{L} / \partial \mu$ is obtained via automatic differentiation, and $\mu$ is updated by the Adam optimizer. After convergence, the final $\mu(z, x)$ map represents the recovered depth-resolved optical attenuation coefficient at each pixel.

The forward model $\mathcal{F}$ is modular: in this work we implement a single-scattering Beer-Lambert model, but the optimization procedure is agnostic to the specific forward physics and requires only that the model is differentiable with respect to $\mu$. Pre-processing steps applied before optimization include subtraction of the noise floor in linear intensity scale and automatic initialization of the system gain offset by median-matching between predicted and target tissue intensities.


\subsection*{Differentiable Forward Model}\label{sec:forward_models}

The detected OCT intensity at depth $z$ can be modeled as the product of the local backscattering amplitude, the round-trip attenuation along the beam path, and depth-dependent sensitivity and confocal functions corrupted by an additive noise floor \cite{faber2004quantitative,vermeer2013depth,cannon2021layer}:
\begin{equation}\label{eq:linear_fwd}
    I(z) = G \cdot \alpha(z) \cdot \mu(z) \cdot \exp\!\Bigl(-2\!\int_0^z \mu(u)\,du\Bigr) \cdot S(z) \cdot H(z) + N(z),
\end{equation}
where $G$ is the system gain, representing the incident irradiance on the sample scaled by detection efficiency, $\alpha(z)$ is the unitless local backscattering fraction~\cite{cannon2021layer}, written as $\alpha(z, x)$ for a B-scan, $\mu(z)$ is the attenuation coefficient [mm$^{-1}$], the exponential term describes round-trip Beer-Lambert attenuation from the surface to depth $z$, $S(z)$ is the dimensionless sensitivity fall-off function in linear intensity scale, and $N(z)$ is the estimated mean depth-dependent noise-floor intensity. The confocal gate $H(z)$ models the depth-dependent collection efficiency of the focusing optics \cite{faber2004quantitative}:
\begin{equation}\label{eq:confocal}
    H(z) = \frac{1}{\left(\dfrac{z - z_f}{z_R}\right)^2 + 1},
\end{equation}
where $z_f$ is the focal plane position and $z_R$ is the Rayleigh range of the sample arm beam. Sensitivity fall-off can occur in both spectral-domain OCT, due to finite spectrometer resolution, and swept-source OCT, due to the finite instantaneous linewidth of the swept laser~\cite{choma2003sensitivity}. The swept-source system used for the phantom studies (based on a Thorlabs SL131090 light source) had negligible sensitivity fall-off over the imaging range spanned by the phantom signal, which was placed close to the zero-path delay. We used $S(z)=1$ for the agar-phantom inversions, applied the measured $S(z)$ during preprocessing for the liquid phantoms, and assumed $S(z)=1$ for the clinical data because scanner-specific sensitivity calibration was unavailable.

Prior to optimization, the noise floor $N(z)$ is subtracted and the sensitivity correction is applied in linear intensity scale \cite{cannon2021layer}:
\begin{equation}\label{eq:nz_correction}
    I_\text{corr}(z) = \frac{I(z) - N(z)}{S(z)} = G \cdot \alpha(z) \cdot \mu(z) \cdot \exp\!\Bigl(-2\!\int_0^z \mu(u)\,du\Bigr) \cdot H(z).
\end{equation}
This expression describes the mean corrected intensity where individual measurements retain residual noise fluctuations that can affect the recovered $\mu$. Taking $10\log_{10}(\cdot)$ of both sides converts the multiplicative forward model into an additive expression in log scale:
\begin{equation}\label{eq:dB_fwd}
    I_\text{dB}(z) = G_\text{dB} + 10 \cdot \log_{10}\!\alpha(z) + 10 \cdot \log_{10}\!\mu(z) + H_\text{dB}(z) - \frac{20}{\ln 10} \cdot \int_0^z \mu(u)\,du,
\end{equation}
where $G_\text{dB} = 10 \cdot \log_{10} G$ is the system gain [dB] and $H_\text{dB}(z) = 10 \cdot \log_{10} H(z)$ is the confocal function correction [dB]. In discrete form, the integral is replaced by a cumulative sum over depth pixels with axial spacing $\Delta z$, where pixel $i$ accumulates attenuation from all pixels above it but not itself:
\begin{equation}\label{eq:dB_fwd_discrete}
    I_\text{dB}(z_i) = G_\text{dB} + 10 \cdot \log_{10}\!\alpha(z_i) + 10 \cdot \log_{10}\!\mu(z_i) + H_\text{dB}(z_i) - \frac{20}{\ln 10} \cdot \sum_{j=0}^{i-1} \mu(z_j) \cdot \Delta z.
\end{equation}

This log-domain formulation is implemented as a differentiable PyTorch module in which $\mu(z, x)$ is the optimizable parameter tensor and, for measured-data attenuation recovery, $\alpha(z, x) = 1$ at every tissue pixel, so that $10 \cdot \log_{10}\alpha = 0$ in Eq.~\eqref{eq:dB_fwd_discrete} and $\alpha$ drops out of the optimization. Because every operation in Eq.~\eqref{eq:dB_fwd_discrete} is differentiable, the gradient $\partial I_\text{dB} / \partial \mu$ is computed exactly via automatic differentiation and propagated to the optimizer (Fig.~\ref{fig:framework}). All $\mu$ values are updated simultaneously. Here, differentiation through the cumulative sum makes the gradient from the intensity error with respect to $\mu(z_j)$ combine its local contribution with contributions from deeper pixels included in the loss. In the clinical second stage, $\mu$ is frozen and $\alpha(z, x)$ is optimized per pixel (Supplementary Note~6).

\subsection*{Inverse Problem Formulation}\label{sec:inverse}

Given a noise-floor-corrected target B-scan $I_\text{target}(z, x)$ in log scale, we seek the attenuation map that minimizes the reconstruction error over tissue pixels:
\begin{equation}\label{eq:loss}
    \mu^*(z, x) = \arg\min_{\mu} \; \frac{1}{|\mathcal{M}|} \sum_{(z,x) \in \mathcal{M}} \bigl[[\mathcal{F}(\mu;\, \boldsymbol{\theta})](z, x) - I_\text{target}(z, x)\bigr]^2,
\end{equation}
where $\mathcal{F}(\mu;\, \boldsymbol{\theta})$ is the forward model prediction (Eq.~\eqref{eq:dB_fwd_discrete}), $\boldsymbol{\theta}$ denotes the fixed system parameters, and $\mathcal{M}$ is the set of tissue pixels whose measured intensity exceeds the noise floor threshold, fixed before optimization. For the agar-phantom and clinical inversions, pixels below this threshold are excluded from both the loss computation and the cumulative attenuation sum in the forward model, preventing noise-dominated regions from influencing the recovered attenuation. To promote spatial smoothness in the recovered attenuation map, the loss is augmented with an anisotropic total variation (TV) penalty \cite{yu2026endtoend}:
\begin{equation}\label{eq:tv}
    \text{TV}(\mu) = \frac{1}{|\mathcal{M}_z|} \sum_{(z,x) \in \mathcal{M}_z} |\mu(z{+}1, x) - \mu(z, x)| + \frac{1}{|\mathcal{M}_x|} \sum_{(z,x) \in \mathcal{M}_x} |\mu(z, x{+}1) - \mu(z, x)|,
\end{equation}
where $\mathcal{M}_z$ and $\mathcal{M}_x$ are the sets of adjacent tissue pixel pairs along the depth and lateral axes, respectively. The total loss is:
\begin{equation}\label{eq:total_loss}
    \mathcal{L}_\text{total}(\mu) = \lambda_\text{MSE} \cdot \mathcal{L}(\mu) + \lambda_\text{TV} \cdot \text{TV}(\mu),
\end{equation}
where the reconstruction term $\mathcal{L}(\mu) = \frac{1}{|\mathcal{M}|} \sum_{(z,x) \in \mathcal{M}} \bigl[\hat{I}_\text{dB}(z,x) - I_\text{target}(z,x)\bigr]^2$ is the masked MSE of Eq.~\eqref{eq:loss}, with $[\mathcal{F}(\mu;\, \boldsymbol{\theta})](z,x) = \hat{I}_\text{dB}(z,x)$ the predicted log-scale intensity at pixel $(z,x)$ from Eq.~\eqref{eq:dB_fwd_discrete}, $I_\text{target}(z,x)$ the corresponding measured intensity [dB], and $|\mathcal{M}|$ the number of tissue pixels above the noise floor. The weights $\lambda_\text{MSE}$ and $\lambda_\text{TV}$ control the relative weighting of the reconstruction and regularization terms. For the agar-phantom inversions, we weight each loss equally $\lambda_\text{MSE} = \lambda_\text{TV}$. The clinical cohort inversions used no TV penalty, whereas the five frames shown in \Cref{fig:clinical} and analyzed in Supplementary Note~8 used $\lambda_\text{MSE} = \lambda_\text{TV}$. The optimizer comparisons specify TV separately because it is disabled in the synthetic scaling sweeps, enabled axially in the measured A-line sweep, and varied in the matched joint and independent fits (Supplementary Note~1). The sensitivity of the recovered attenuation to $\lambda_\text{TV}$ is characterized in the sensitivity study (Supplementary Note~3).

The attenuation map is optimized using the Adam optimizer. At each iteration, gradients $\partial \mathcal{L} / \partial \mu$ are computed via backpropagation through the forward model and applied to update $\mu$. After each optimizer step, $\mu$ is projected onto the feasible interval $[\mu_\text{min}, \mu_\text{max}]$ and, for the agar-phantom and clinical inversions, noise-floor pixels are re-zeroed. In these inversions, the optimizer tracks the best reconstruction loss throughout training and restores the corresponding $\mu$ at convergence.

The system gain $G_\text{dB}$ is initialized prior to optimization by running the forward model with the initial $\mu$ and computing the median offset between predicted and target tissue intensities. This closed-form initialization places the predicted B-scan in the correct intensity range before optimization begins. For the agar-phantom and clinical inversions, $G_\text{dB}$ is recomputed at each iteration as the MSE-optimal additive offset:
\begin{equation}\label{eq:gain_reopt}
    G_\text{dB}^* = \frac{1}{|\mathcal{M}|} \sum_{(z,x) \in \mathcal{M}} \bigl[I_\text{target}(z,x) - \mathcal{F}(\mu;\, \boldsymbol{\theta})\big|_{G_\text{dB}=0}(z,x)\bigr],
\end{equation}
where $\mathcal{F}(\mu;\, \boldsymbol{\theta})\big|_{G_\text{dB}=0}$ is the forward model prediction with gain set to zero. This per-iteration recomputation minimizes the residual offset for the current $\mu$; it does not independently calibrate the attenuation scale.

\subsection*{Computational Verification Protocol}\label{sec:comp_verification}

To verify numerical correctness of the forward and inverse models, we conducted two complementary tests. First, the forward model was evaluated on 500 synthetic A-lines, each containing 2 to 5 homogeneous layers with randomized attenuation coefficients drawn from $[0.1, 15.0]$~mm$^{-1}$. With the confocal function disabled, system gain set to unity, and $\alpha = 1$, the forward model output was compared pixel-by-pixel against the analytical Beer-Lambert solution computed independently. Agreement was quantified by the coefficient of determination ($R^2$) and the maximum absolute error across all pixels.

Second, the inverse optimization was evaluated on 500 synthetic B-scans with spatially varying, per-pixel ground-truth attenuation. Each B-scan was generated by the forward model with known $\mu$, $\alpha$, confocal parameters, and system gain. The attenuation map was then scrambled to random initial values and recovered by the optimizer over 30,000 iterations, with $\alpha$ and $G_\text{dB}$ frozen at their true values to isolate the recovery of $\mu$.

To assess scaling with coupled parameters, the same attenuation field was sampled within a 0.16~mm tissue band at $K=25,50,100,200,500,1{,}000,2{,}000,$ and $5{,}000$ depths in one A-line. Axial spacing was $\Delta z=0.16/K$~mm; focal position and Rayleigh range in pixels were rescaled to preserve the physical confocal geometry. Random Search~\cite{solis1981minimization}, CMA-ES~\cite{hansen2001completely}, Adam~\cite{kingma2014adam}, and L-BFGS-B~\cite{byrd1995limited} shared the target, masks, fixed gain and backscattering fraction, OAC bounds of $[0.01,20]$~mm$^{-1}$, and eight paired initializations drawn uniformly from $[0.01,10]$~mm$^{-1}$. Each optimizer used a nominal budget of 20,000 iterations (Adam, Random Search, and L-BFGS-B) or objective evaluations (CMA-ES). Recovery accuracy was quantified by the mean absolute percentage error (MAPE) between recovered and ground-truth tissue attenuation:
\begin{equation}\label{eq:mape}
    \text{MAPE} = \frac{1}{|\mathcal{T}|} \sum_{(z,x) \in \mathcal{T}} \frac{|\hat{\mu}(z,x) - \mu_\text{true}(z,x)|}{\mu_\text{true}(z,x)} \times 100\%,
\end{equation}
where $\mathcal{T}$ denotes the set of tissue pixels. The MAPE is evaluated on the recovered attenuation coefficient rather than on image intensity. Wall-clock time per evaluation (ms/eval) was also recorded to quantify per-iteration compute cost. Per-evaluation cost differs across methods because each evaluation carries different work: Random Search runs only a forward pass; Adam adds a backward pass to compute gradients; L-BFGS-B additionally maintains a running history of past gradients to build its quasi-Newton search direction, and that bookkeeping cost grows as the parameter count increases; CMA-ES on top of evaluation must update and occasionally re-decompose a covariance matrix over the parameters, so its per-evaluation cost climbs steeply with problem size and dominates at large problem scales. Supplementary Note~1 describes the synthetic B-scan sweep, with 40 parameters per A-line, and the measured-data comparisons.

\subsection*{Physical Phantom Validation}\label{sec:methods_phantom}

For the liquid comparison, five intralipid concentrations (0.5\%, 1.0\%, 1.5\%, 2.0\%, and 3.0\%) were imaged at two focal positions (H and L; \Cref{fig:phantom_boxplot}a--c), with 512 B-scans of $500\times500$ pixels per acquisition. The system used a 1300~nm source with 95~nm bandwidth and an LSM03 scan lens (NA $\approx0.05$), and axial sampling in water was 7.35~\textmu m. Despeckled 16-bit intensities were converted as $I_\mathrm{dB}=I_\mathrm{raw}\times150/65535$. Linear intensities were corrected as $I_\text{corr,liq}(z)=[I(z)-N(z)]/[S(z) \cdot H(z)]$ using the measured noise and sensitivity curves and calibrated confocal function ($z_R=0.4315$~mm, with a separate focal position for each B-scan). Both CF and DiffOCT were fitted to these noise-, sensitivity-, and confocal-corrected liquid-phantom intensities. Fitting extended from the start of each A-line's 100-pixel CF window to its last valid depth. Only finite, positive corrected measurements entered the fixed loss mask; excluded observations within this depth interval remained in the cumulative attenuation path.

For each A-line, CF supplied a constant $\mu_\mathrm{CF}$, fitted as a log-linear decay over a 100-pixel depth window (0.735~mm), and the natural-log intensity at the window start, $b_s$. Its equivalent DiffOCT gain, $G_\mathrm{dB}=10b_s/\ln(10)-10\log_{10}\mu_\mathrm{CF}$, was held fixed throughout optimization. Thus, gain calibration uses both the CF intercept and attenuation estimate. With $\alpha=1$, maps were initialized uniformly over $[0,5]$~mm$^{-1}$ and optimized with Adam for 15,000 iterations, using a cosine learning-rate decay from $10^{-3}$ to $10^{-5}$, bounds of $[10^{-6},20]$~mm$^{-1}$, and $\lambda_\mathrm{MSE}=\lambda_\mathrm{TV}=1000$ with axial and lateral TV.

Additionally, five different in vitro phantoms were fabricated at different intralipid-in-agar concentrations (1.0\%, 1.5\%, 2.0\%, 3.0\%, and 5.0\%), each imaged at two focal positions, and were used for visualizing the optimization workflow (Supplementary Figure~2b) and for the identifiability and noise robustness studies (\Cref{fig:ident_robust}). Each phantom was imaged with a wavelength-swept source OCT system with negligible sensitivity fall-off (central wavelength 1300~nm, bandwidth 90~nm, described in detail elsewhere~\cite{cannon2021layer}), with an LSM04 scan lens (NA = 0.0315), each acquisition producing a volumetric stack of 512 B-scans at $500 \times 500$ pixels per frame (500 depth samples $\times$ 500 A-lines). Intensity $I_\text{raw}$ data were stored as 16-bit TIFF files in log-intensity scale, with the mapping $I_\text{dB} = I_\text{raw} \times (130 / 65535)$ yielding values in the 0--130~dB.

The two tomograms from each phantom were registered in three dimensions via a fast Fourier transform to automatically determine the axial shift $\delta z$ together with small lateral shifts arising from deviation from pure axial translation. Speckle was reduced by Gaussian spatial filtering with $1/e^2$-diameters of $7 \times 15$~pixels along the fast-scan and slow-scan axes, respectively, applied to all volumes prior to optimization.

The paired tomograms and the recovered axial shift were used to estimate the focal plane position $z_f$ and effective Rayleigh range $z_R$ via logarithmic-intensity matching~\cite{dwork2016automated}. Specifically, the optimal parameters were found by minimizing the mean-squared difference between the two confocal-corrected tomograms:
\begin{equation}\label{eq:confocal_est}
\begin{split}
    \{z_f^*, z_R^*\} = \arg\min_{z_f,\, z_R}
    \frac{1}{N}\sum_{z,x,y}
    \bigl\lvert
        I^{(1)}_\text{dB}(z,x,y) - H_\text{dB}(z;\, z_f,\, z_R) \\
      - \bigl[I^{(2)}_\text{dB}(z,x,y)
              - H_\text{dB}(z - \delta z;\, z_f,\, z_R)\bigr]
    \bigr\rvert^2,
\end{split}
\end{equation}
where $I^{(1)}$ and $I^{(2)}$ are the two registered tomograms from the same phantom, $N$ is the number of voxels, and $H_\text{dB}$ is the confocal gate (Eq.~\eqref{eq:confocal}) expressed in log scale.

Independent reference attenuation coefficients were obtained by conventional curve-fitting (CF) slope analysis performed in MATLAB on the same phantom volumes. After noise-floor subtraction and correction for the calibrated confocal function, a linear model $I_\text{ln}(z) = b_0 + b_1 \cdot z$, with $z$ the depth-pixel index, was fitted to the natural-logarithmic intensity of 40 depth pixels (11--50 below the surface peak) within the thresholded tissue mask. The attenuation coefficient was then computed as $\mu_\text{CF} = -b_1/(2\delta_z)$, where $\delta_z$ is the axial pixel size in water and the factor of $1/2$ accounts for the double-pass geometry.

Each of the 512 B-scans per acquisition was processed independently using the single-scattering forward model (Eq.~\eqref{eq:dB_fwd_discrete}) with noise floor subtraction (Eq.~\eqref{eq:nz_correction}) and per-iteration gain reoptimization (Eq.~\eqref{eq:gain_reopt}). Recovered attenuation coefficients were evaluated exclusively over tissue pixels defined by a binary segmentation mask derived from independent MATLAB ground-truth labeling, ensuring that reported $\mu$ statistics excluded noise-floor-dominated regions where the optimizer assigns zero attenuation.

\subsection*{Identifiability}\label{sec:methods_identifiability}

The identifiability study tested whether the recovered attenuation is independent of optimization randomness and initialization. In the first experiment, each acquisition was optimized 20 times with different random seeds controlling the initial $\mu$ values, while keeping all other parameters fixed. The coefficient of variation (CoV) of the recovered tissue $\mu$ across seeds was used to quantify reproducibility. In the second experiment, the initialization range was varied across 10 conditions spanning narrow ($[1, 1]$~mm$^{-1}$), wide ($[0, 20]$~mm$^{-1}$), and biased ($[7, 15]$~mm$^{-1}$) ranges, with 5 seeds per condition per acquisition. Agreement of the converged $\mu$ across initialization conditions was assessed by the CoV and by the range of recovered tissue means.

\subsection*{Noise Robustness}\label{sec:methods_robustness}

The robustness study quantified how additive noise degrades attenuation recovery. Zero-mean, circular Gaussian noise $\sim\mathcal{N}(0,\sigma_n)$ was added to the complex-valued data at controlled signal-to-noise ratios (SNR). The noise variance $\sigma_n^2$ to achieve a target SNR was calculated as
\begin{equation}
\sigma_n^2 = \bar{I}_n\left( \frac{R - R'}{1+R'} \right),
\end{equation}
where $R$ and $R'$ are the original and target SNRs, defined with the rigorous definition $R=(\bar{I}_s - \bar{I}_n)/\bar{I}_n$~\cite{uribe2020noise}, where $\bar{I}_s$ and $\bar{I}_n$ represent the average intensity at a signal-dominated region (a slab covering 10 depth pixels below the phantom surface) and a noise-dominated region (the air region above the phantom surface), respectively. The noise injection protocol was applied to one B-scan from each of the 10 phantom acquisitions, with the noiseless recovery serving as the baseline reference. Degraded-SNR cases varied from $-3$~dB up to each phantom's original SNR in 1~dB steps. Degradation was quantified by the change in recovered tissue $\mu$ relative to the noiseless condition. For reference, SNR $= 20$~dB corresponds to a signal intensity two orders of magnitude above the noise floor, while SNR $= 0$~dB corresponds to noise equal to the signal.

\subsection*{Application to Clinical IV-OCT}\label{sec:methods_clinical}

The clinical demonstration used 29{,}269 pre-IVL intravascular OCT frames across 128 patients from the Shockwave Disrupt CAD-I through CAD-IV clinical trial cohorts~\cite{ali2017optical,ali2019safety,kereiakes2023intravascular,saito2022intravascular}. IV-OCT images were stored as $500 \times 500$ pixel Cartesian cross-sections. The five representative frames displayed in~\Cref{fig:clinical} were acquired with an ILUMIEN OPTIS system and a C7-XR Dragonfly catheter (Abbott Vascular).
         
Each Cartesian cross-section was converted to a polar representation ($250~\text{depth} \times 500~\text{A-lines}$) by extracting radial A-lines from the catheter center outward. The Rayleigh range $z_R$ and focal plane position $z_f$ were set to 31~pixels (${\approx}0.59$~mm) and 55~pixels (${\approx}1.06$~mm) respectively, adopting the measured confocal parameters reported by Gargesha et al.\ for the C7-XR / Dragonfly catheter family~\cite{gargesha2015parameter}. The axial pixel size of $0.01932$~mm/pixel follows from the ILUMIEN OPTIS A-scan range of $4.83$~mm in contrast over 250 depth samples per the device's 510(k) summary (FDA K141453, \url{https://www.accessdata.fda.gov/CDRH510K/K141453.pdf}); the numerical reconstruction used $0.020$~mm/pixel. Per-pixel attenuation maps $\mu(z, x)$ were recovered over 25{,}000 iterations with the backscattering fraction $\alpha$ held fixed at unity (Stage~1). Our optimized $\mu$ maps were then frozen and a per-pixel backscattering map $\alpha \in [10^{-10}, 10^{-2}]$ (bounds informed by supplementary of~\cite{cannon2021layer}) was optimized over an additional 25{,}000 iterations (Stage~2) to decouple per-pixel brightness from cumulative attenuation. Additional details of this two-stage procedure are given in Supplementary Note~6. The noise floor was subtracted in linear intensity scale and the system gain was recomputed at each iteration. Rendering quality was assessed by the peak signal-to-noise ratio (PSNR) between the forward model prediction and the noise-floor-corrected target image, computed in log-intensity scale over tissue pixels as

\begin{equation}
\mathrm{PSNR} = 10 \log_{10}\!\left(\frac{D^{2}}{\mathrm{MSE}}\right),
\label{eq:psnr}
\end{equation}
where $\mathrm{MSE} = \frac{1}{|\mathcal{T}|}\sum_{(z,x)\in\mathcal{T}}\big(\hat{I}_{\mathrm{dB}}(z,x) - I_{\mathrm{dB}}(z,x)\big)^{2}$ is the mean squared residual between the rendered ($\hat{I}_{\mathrm{dB}}$) and target ($I_{\mathrm{dB}}$) log-scale images evaluated only on the tissue pixel set $\mathcal{T}$ defined by the noise-floor mask, and $D = \max_{\mathcal{T}} I_{\mathrm{dB}} - \min_{\mathcal{T}} I_{\mathrm{dB}}$ is the log-scale dynamic range of the tissue signal in the target. Target images were encoded with a full log-scale of 130~dB and a tissue mask threshold of 15~dB, giving a tissue dynamic range $D \approx 115$~dB that is essentially constant across frames. Higher PSNR values indicate that the rendered image matches the target more closely relative to the available tissue signal range; because the denominator grows with MSE, a $k$-fold reduction in MSE increases PSNR by $10\log_{10} k$~dB.

In addition to PSNR, tissue-coverage agreement between the rendered and target B-scans was quantified by the Dice overlap of their binarized tissue masks. Both images were binarized at the 15~dB noise floor threshold that the optimizer uses to define tissue pixels, and the Dice coefficient was computed over the full Cartesian frame as $\mathrm{Dice} = 2|A \cap B| / (|A| + |B|)$, where $A$ and $B$ are the sets of pixels above threshold in the target and rendered images respectively. This metric captures whether the two images agree on which pixels carry signal above the noise floor, complementing the intensity-sensitive PSNR.

\subsection*{Statistical Analysis}\label{sec:stats}

All quantitative analyses were performed in Python (v3.10) using NumPy (v1.24), SciPy (v1.11), and Matplotlib (v3.7). The study is descriptive throughout: no inferential hypothesis tests were applied, as this work compares recovered properties to known references and reports per-pixel error distributions rather than testing population effects. Continuous quantities computed across frames, phantoms, or random seeds are reported as mean $\pm$ standard deviation unless otherwise noted, and per-pixel distributions are summarized by median, mean absolute error, and the fraction of pixels within fixed dB tolerance envelopes ($\pm 3$, $\pm 10$, $\pm 20$~dB). For the phantom validation, linear regression of recovered tissue attenuation against intralipid concentration was fit by ordinary least squares and reported as slope, intercept, and the coefficient of determination $R^2$. For the identifiability study, the within-acquisition coefficient of variation (CoV) of recovered tissue-mean $\mu$ was computed as the standard deviation divided by the mean across the perturbation set and reported as a percentage. Initialization-range CoV was computed after averaging five seeds per condition, using the seven selected phantom conditions or all ten clinical conditions (\Cref{fig:ident_robust}c,d and Supplementary Figure~8b). For the noise robustness study, recovery degradation was quantified as the absolute relative change in recovered tissue $\mu$ versus the noiseless baseline ($|\Delta\mu|/\mu_\text{baseline}$), and spatial agreement was tracked as the pixel-wise $R^2$ of recovered $\mu$ versus the noiseless baseline at each SNR.

For \Cref{fig:optimizer_baseline}, attenuation MAPE (Eq.~\eqref{eq:mape}) was computed separately for each run against the known tissue attenuation. Runtime per objective evaluation was total solver wall time divided by the actual number of evaluations, including solver overhead. Points and error bars report the arithmetic mean and standard deviation across $N=8$ paired initializations at each problem size. These repeats measure numerical variability on the same synthetic target.

For \Cref{fig:phantom_boxplot}f, DiffOCT attenuation was first averaged over valid pixels within each 100-pixel CF window (0.735~mm), then equally over paired A-lines to obtain one mean per B-scan. CF used the same A-lines; lines with nonfinite or nonpositive CF estimates, empty valid windows, or fallback gain were excluded from both summaries. Each acquisition contributed 512 frame means. Violin densities and black mean markers use values within $[Q_1-1.5\,\mathrm{IQR},Q_3+1.5\,\mathrm{IQR}]$, applied separately to each method, concentration, and focal position; $\mathrm{IQR}=Q_3-Q_1$. This filtering affects only the display (486--512 retained frames per violin). Regression fits use all frame means, averaged to five equally weighted concentration means for each focal position; the displayed pooled CF line uses ten acquisition means. $R^2$ measures concentration linearity rather than agreement between methods. Attenuation bias is the difference between unfiltered DiffOCT and CF volume means. In \Cref{fig:phantom_boxplot}g, intensity RMSE is the square root of the mean squared dB residual over the same valid CF-window pixels for both methods.


\section*{Ethics Statement}
The physical phantom experiments involved no human subjects, animal subjects, or biological tissue and required no ethical approval. The clinical intravascular OCT data were acquired as part of the Shockwave Disrupt CAD-I through CAD-IV (ClinicalTrials.gov Identifiers NCT02650128, NCT03328949, NCT03595176, and NCT04151628) study~\cite{ali2017optical,ali2019safety,kereiakes2023intravascular,saito2022intravascular}, sponsored by Shockwave Medical, Inc. All necessary patient consents were secured for the original studies. Ethical approval for secondary use of this dataset was obtained from the Institutional Review Board at Mass General Brigham under protocol IRB~\#2022P001679 (Personalized lesion modification optimizes atherosclerosis intervention); the approval is active and exempt. All protected health information was de-identified according to the Health Insurance Portability and Accountability Act (HIPAA) requirements, and identifiers such as names and birth dates were removed from structured data sources.

\section*{Data Availability}
The clinical intravascular OCT data analyzed in this study were provided by Shockwave Medical, Inc. and are not publicly available because their sharing is restricted by the data use agreement with Shockwave Medical, Inc. The intralipid phantom OCT data generated and analyzed in this study are available from the corresponding author, Ajay Manicka (ajay\_m@mit.edu), upon reasonable request. Requests will normally be answered within 3 weeks. The source data for Figures 1--5 are provided in Supplementary Data 1. Additional material is provided as Supplementary Information.

\section*{Code Availability}
All code used in this study, including the code to generate the synthetic phantoms, will be released on GitHub and archived on Zenodo with a DOI upon publication. The code is GPU-accelerated but also CPU-compatible.

\section*{Acknowledgements}
The authors would like to thank Taylor Cannon and Louis Leger for their helpful discussions throughout the duration of this article. We also thank Shockwave Medical Inc. for providing the rich dataset used in this study. We acknowledge Albert Swiston and MIT Supercloud for their support and computing resources. Figures were made using Biorender software and Python visualization libraries.

\section*{Funding} This work was supported by a grant to E.R.E and F.N. from the National Institutes of Health (R01 HL161069). F.N. graciously acknowledges career funding from American Heart Association (25CDA1452999).

\section*{Competing Interest} The authors declare no competing interests.

\section*{Author Information}
A.M. conceived, designed, and executed the project. K.K. provided technical support for the inverse modeling design. S.R-L. conducted experimental testing and validation on intralipid phantoms. M.A. provided substantial editorial contributions to the manuscript and helped design studies. S.A. helped develop figures and contributed to the discussion of results. S.R-C helped perform analyses and contributed to the discussion of results. N.U-P. and B.E.B. provided guidance on OCT physics and forward model design. F.N. and E.R.E. supervised and advised A.M. throughout all stages of the project. All authors contributed to the review and editing of the manuscript.

\newpage
\bibliography{bibliography}

\clearpage
\section*{Figures}

\begin{center}
\textbf{Figure~\ref{fig:optimizer_baseline}}\\[6pt]
\includegraphics[width=\linewidth]{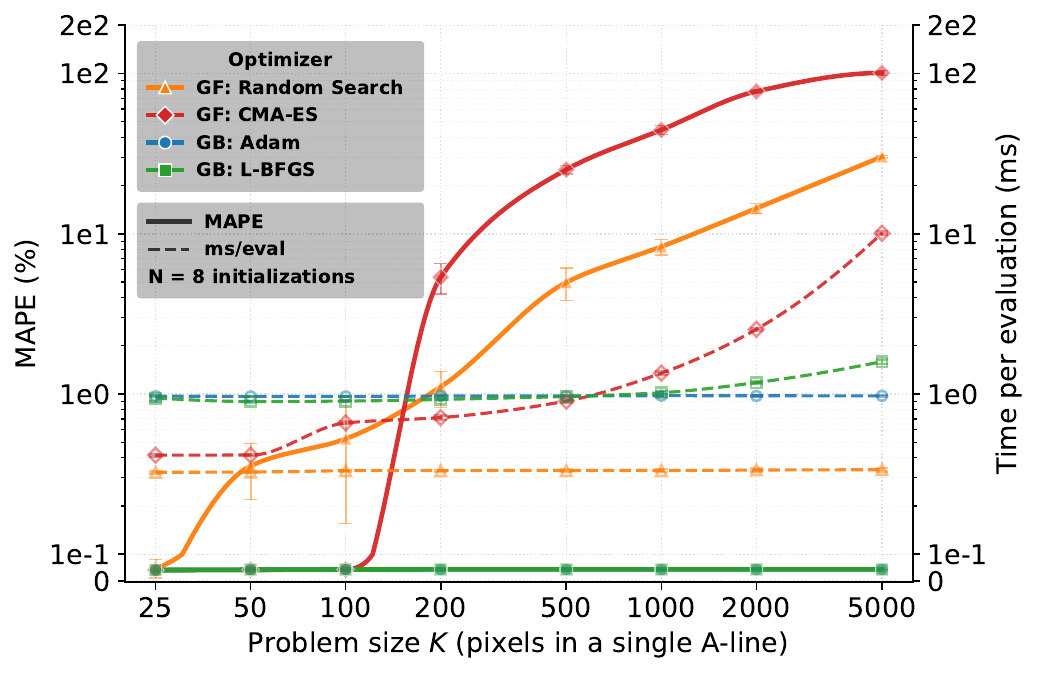}
\end{center}
\clearpage

\begin{center}
\textbf{Figure~\ref{fig:verification}}\\[6pt]
\includegraphics[width=\linewidth]{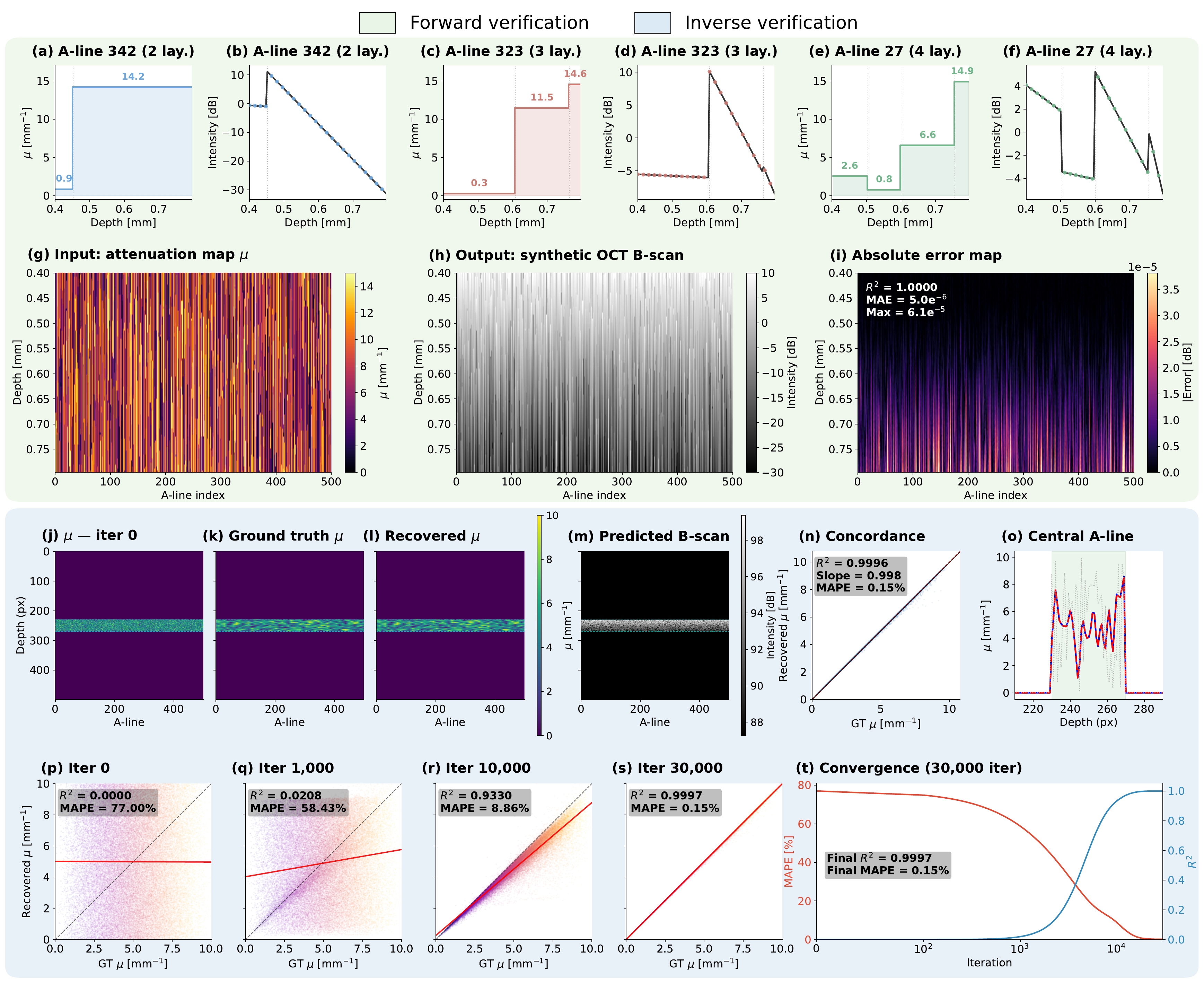}
\end{center}
\clearpage

\begin{center}
\textbf{Figure~\ref{fig:phantom_boxplot}}\\[6pt]
\includegraphics[width=\linewidth]{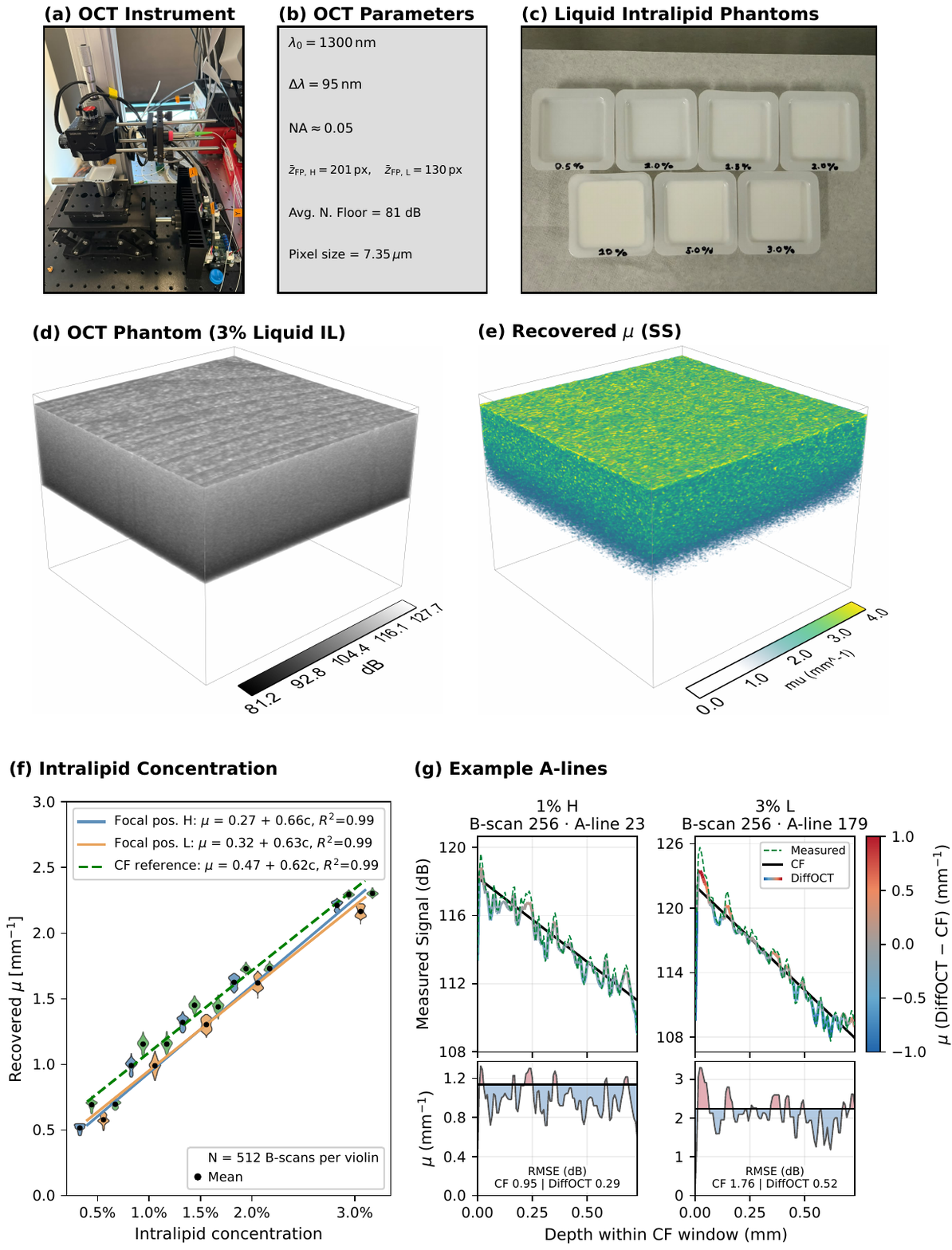}
\end{center}
\clearpage

\begin{center}
\textbf{Figure~\ref{fig:ident_robust}}\\[6pt]
\includegraphics[width=\linewidth]{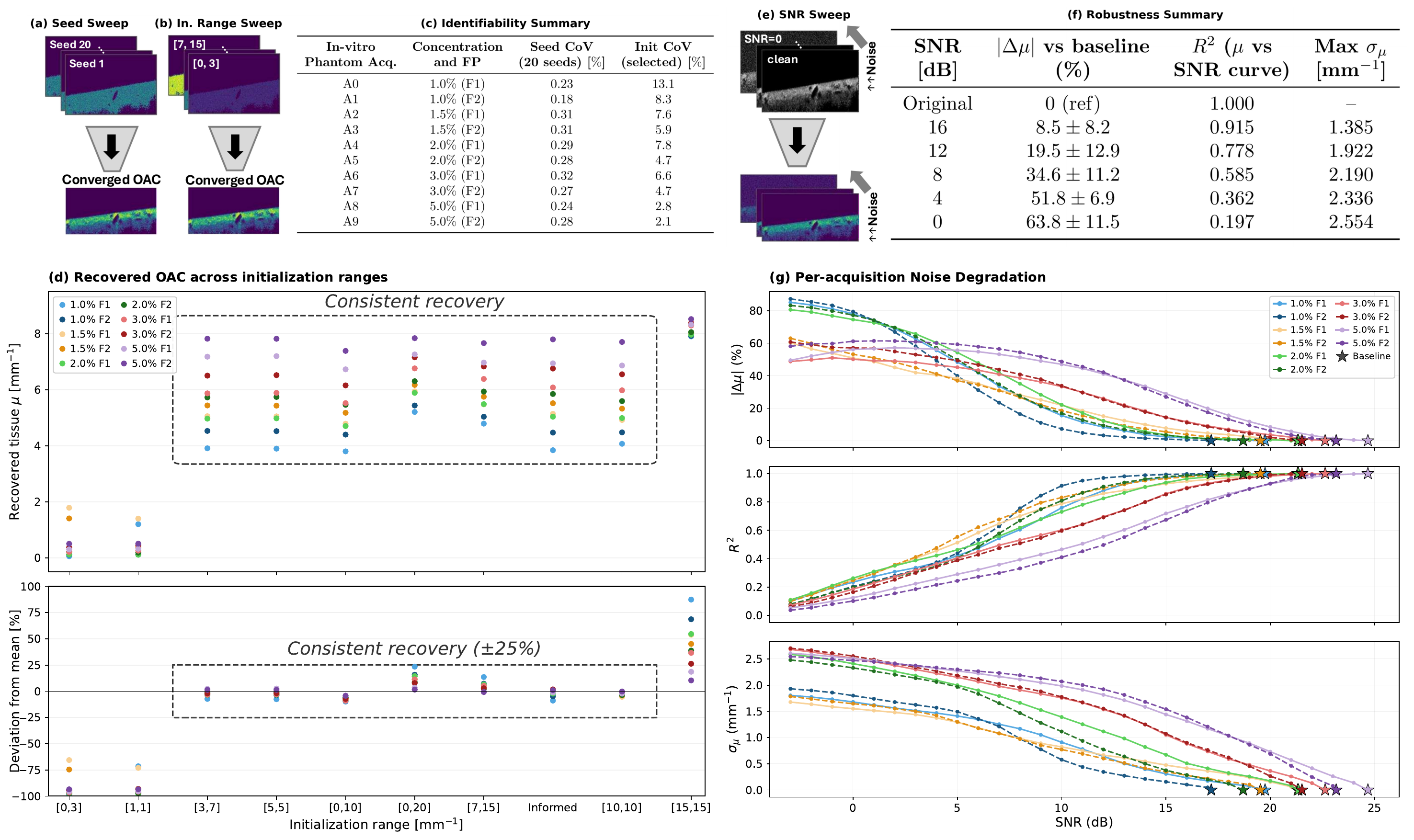}
\end{center}
\clearpage

\begin{center}
\textbf{Figure~\ref{fig:clinical}}\\[6pt]
\includegraphics[width=\linewidth]{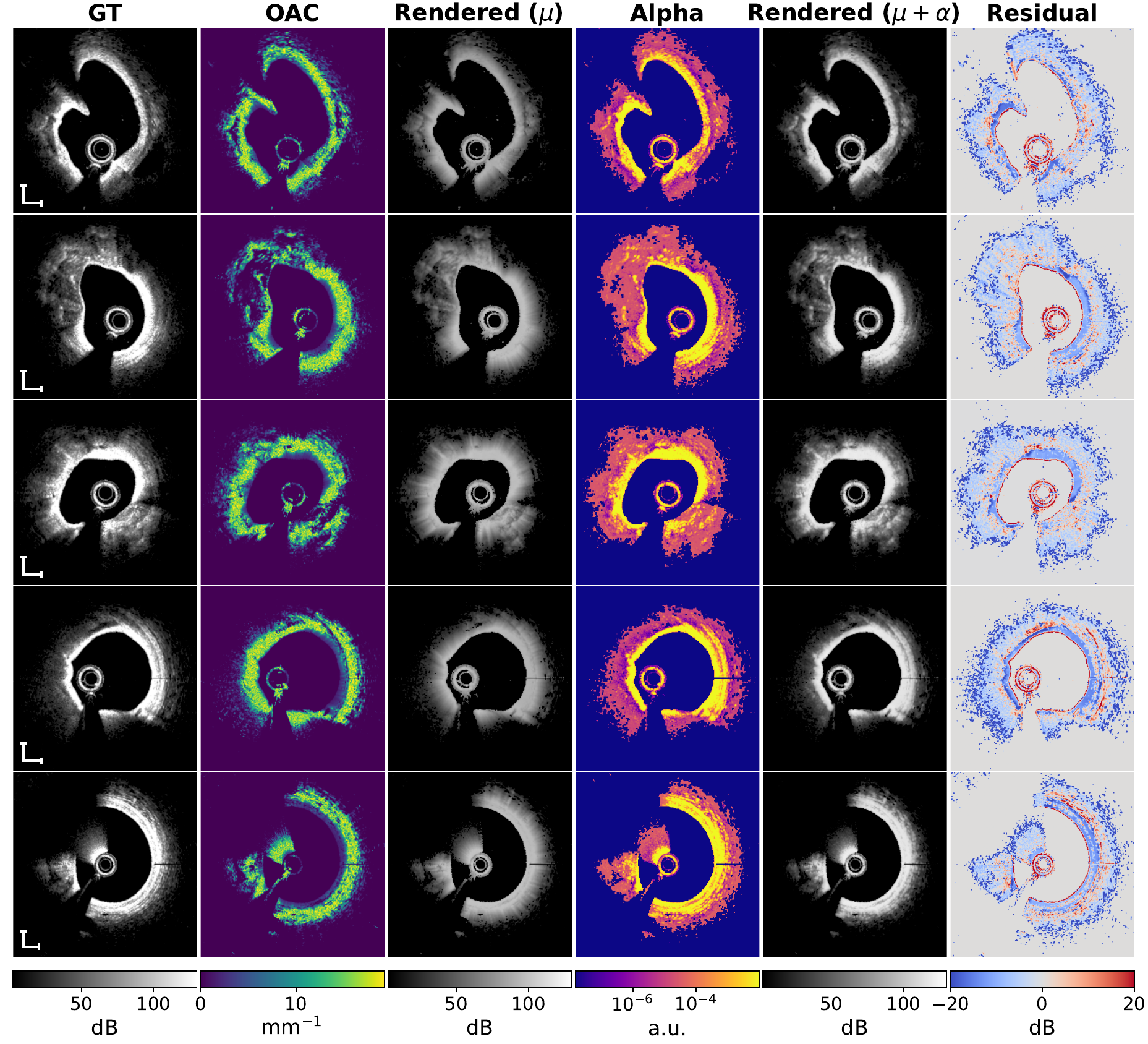}
\end{center}
\clearpage

\begin{center}
\textbf{Figure~\ref{fig:framework}}\\[6pt]
\includegraphics[width=\linewidth]{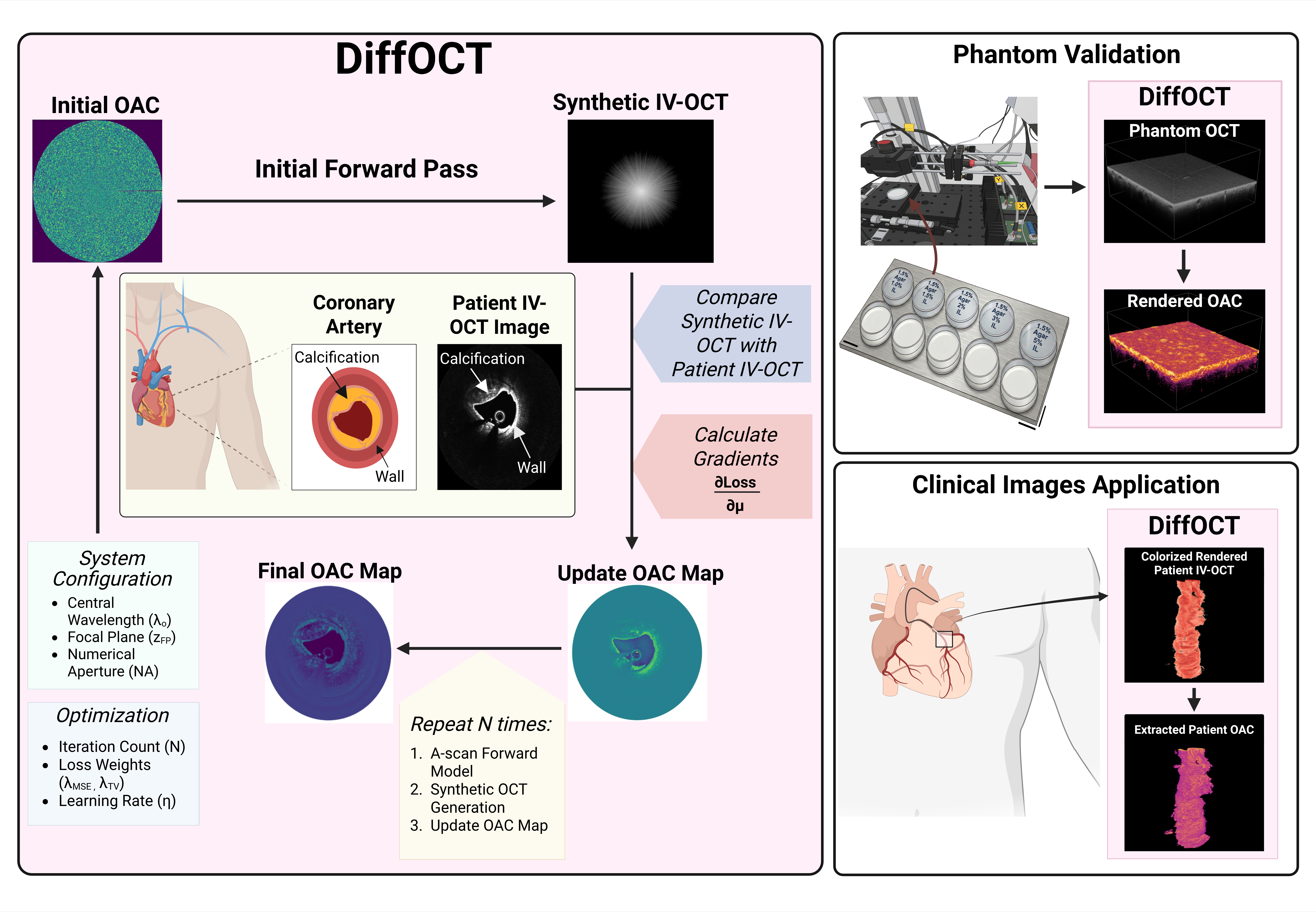}
\end{center}

\clearpage
\section*{Figure Captions}

{\captionsetup{type=figure}
\caption{\textbf{Gradient-based optimization maintains attenuation recovery accuracy as the number of coupled depth parameters increases.} Solid curves show attenuation MAPE against known ground truth (\Cref{eq:mape}); dashed curves show wall time per objective evaluation. Problem size $K$ counts depth parameters in one synthetic A-line, spanning eight sizes from 25 to 5,000 within a fixed 0.16~mm tissue band. Random Search, CMA-ES, Adam, and L-BFGS-B share the forward model, bounds, and initial values, with TV disabled and solver-specific optimization budgets. GF and GB denote gradient-free and gradient-based optimizers, respectively. Points and error bars denote mean $\pm$ one standard deviation across $N=8$ paired initializations; curves interpolate the means. Timing includes solver overhead and is normalized by actual evaluation counts on NVIDIA Tesla V100-PCIE-32GB GPUs.}
\label{fig:optimizer_baseline}}
\bigskip

{\captionsetup{type=figure}
\caption{\textbf{Forward and inverse model verification.} (\textbf{a})--(\textbf{f})~Representative A-line profiles for 2-layer (\textbf{a},\textbf{b}), 3-layer (\textbf{c},\textbf{d}), and 4-layer (\textbf{e},\textbf{f}) configurations, showing input attenuation $\mu$ (left) and corresponding dB intensity (right); red dashed lines indicate the analytical Beer-Lambert solution. (\textbf{g})~Input attenuation map $\mu$ for a representative B-scan (500~A-lines, 2--5 layers each). (\textbf{h})~Synthetic OCT B-scan generated by the forward model. (\textbf{i})~Absolute error map between forward model output and analytical solution ($R^2 = 1.0000$, MAE $= 5.0 \times 10^{-6}$~dB, max error $= 6.1 \times 10^{-5}$~dB). (\textbf{j})--(\textbf{o})~Inverse model verification on a representative synthetic B-scan: (\textbf{j})~scrambled initial $\mu$; (\textbf{k})~ground-truth $\mu$; (\textbf{l})~recovered $\mu$ after 30,000 iterations; (\textbf{m})~predicted B-scan at convergence; (\textbf{n})~concordance scatter plot ($R^2 = 0.9996$, slope $= 0.998$, MAPE $= 0.15\%$); (\textbf{o})~central A-line depth profile: scrambled \emph{initial} $\mu$ before optimization (gray dotted), ground-truth $\mu$ (blue dashed), and recovered $\mu$ (red solid). The gray dotted curve is the random starting point, so the comparison of interest is recovered versus ground truth, which overlap to within the line width. (\textbf{p})--(\textbf{s})~Concordance scatter plots of recovered versus ground-truth $\mu$ across all 500 computational phantoms at four optimization snapshots. In each panel the red solid line is the ordinary least-squares fit. Snapshots are taken at iteration 0 ($R^2 < 0.0001$, MAPE $= 77.0\%$), iteration 1{,}000 ($R^2 = 0.02$, MAPE $= 58.4\%$), iteration 10{,}000 ($R^2 = 0.93$, MAPE $= 8.9\%$), and iteration 30{,}000 ($R^2 = 0.9997$, MAPE $= 0.15\%$). (\textbf{t})~Convergence of MAPE (red, left axis) and concordance $R^2$ (blue, right axis) over 30{,}000 iterations on a logarithmic iteration axis, reaching final values of $R^2 = 0.9997$ and MAPE $= 0.15\%$.}\label{fig:verification}}
\bigskip

{\captionsetup{type=figure}
\caption{\textbf{Attenuation recovery in liquid intralipid phantoms.} (\textbf{a})~Swept-source OCT instrument used for phantom imaging. (\textbf{b})~OCT system and image processing parameters. (\textbf{c})~Liquid intralipid phantoms at five concentrations (0.5--3.0\%), each imaged at two focal positions (H and L). (\textbf{d},\textbf{e})~Measured intensity and recovered single-scattering attenuation for the 3.0\% H acquisition, displayed over the same 1.65~mm region from each CF-window start. (\textbf{f})~Distributions of B-scan mean attenuation within CF windows (N = 512 frames). Blue and orange denote DiffOCT at H and L; adjacent green violins denote CF. Black markers are retained-frame means; regression lines use unfiltered volume means, with the dashed CF line pooling H and L. (\textbf{g})~Two example A-lines from B-scan 256 show corrected measured intensity, CF fit, and DiffOCT prediction (top), with corresponding attenuation profiles (bottom). Prediction color and profile shading indicate local $\mu_\mathrm{DiffOCT}-\mu_\mathrm{CF}$; annotations report intensity RMSE within the CF window.}\label{fig:phantom_boxplot}}
\bigskip

{\captionsetup{type=figure}
\caption{\textbf{Identifiability and noise robustness.} (\textbf{a})~Seed sweep experimental design: each acquisition is optimized 20 times with different random initializations, yielding consistent tissue-mean OAC. (\textbf{b})~Initialization range sweep: the $\mu$ initialization interval is varied across 10 conditions with 5 seeds each. (\textbf{c})~Identifiability summary table for all 10 acquisitions (concentration and focal position indicated), reporting per-acquisition seed CoV across the 20 random seeds and per-acquisition init CoV across the 7 initialization ranges satisfying the consistency criterion in~(\textbf{d}). \textbf{d}~Recovered tissue $\mu$ across the 10 initialization ranges for all 10 acquisitions (top), and the corresponding deviation from each acquisition's mean (bottom); the dashed boxes mark the 7 conditions for which every acquisition lies within $\pm25\%$ of its mean across these conditions, after averaging five seeds per condition. The three conditions $[0,3]$, $[1,1]$, and $[15,15]$~mm$^{-1}$ do not satisfy this criterion. (\textbf{e})~Noise robustness experimental design: Gaussian noise is injected at controlled signal-to-noise ratios (SNR) into phantom B-scans, which are then optimized to recover OAC maps. (\textbf{f})~Noise robustness summary table reporting $|\Delta\mu|$ from the noiseless baseline, $R^2$ of recovered $\mu$ versus its noiseless baseline, and the maximum spatial standard deviation $\sigma_\mu$ of recovered tissue $\mu$, at six SNR levels (Original, 16, 12, 8, 4, and 0~dB). Values of $|\Delta\mu|$ are mean $\pm$ SD across $N=10$ acquisitions. (\textbf{g})~Per-acquisition curves of $|\Delta\mu|$, $R^2$, and $\sigma_\mu$ versus SNR for all 10 phantoms. Degradation remains marginal over the first 4--5~dB of added noise below each phantom's baseline SNR and diverges sharply as SNR falls further.}\label{fig:ident_robust}}
\bigskip

{\captionsetup{type=figure}
\caption{\textbf{Clinical IV-OCT attenuation and backscattering fraction estimation.} The framework was applied without modification to all 29,269 frames over the 128 patients. We show 5 representative cross-sections from a single pullback here. The white scale bars delineate 500$\mu$m. Columns show, left to right: ground-truth IV-OCT (GT, dB); recovered OAC map ($\mu$, mm$^{-1}$); Stage~1 rendered image from OAC map with $\alpha = 1$ (dB); recovered per-pixel backscattering map ($\alpha$); Stage~2 rendered image from the sequential $\mu + \alpha$ model (dB); and the final pixel-wise residual (rendered $-$ ground-truth, dB). Recovered attenuation maps are spatially resolved across all frames and track the tissue architecture visible in the raw images, including the vessel wall, lumen boundary, and calcified deposits. Across the full 128-patient cohort, Stage~1 rendered images reproduce the target geometry with Dice $= 0.957 \pm 0.007$ at the 15~dB noise floor and  PSNR $= 14.83 \pm 0.99$~dB. Optimizing $\alpha$ with $\mu$ frozen (Stage~2) decouples per-pixel brightness from cumulative attenuation and raises the cohort-wide PSNR to $22.63 \pm 0.74$~dB (Dice $= 0.954 \pm 0.009$). Cohort values are mean $\pm$ SD across $N=29{,}269$ frames.}\label{fig:clinical}}
\bigskip

{\captionsetup{type=figure}
\caption{\textbf{Framework overview.} A measured OCT B-scan is compared against the output of a differentiable physics-based forward model. Automatic differentiation provides gradients of the reconstruction loss with respect to the per-pixel attenuation map $\mu(z,x)$, which is iteratively updated via gradient-based optimization until the predicted image matches the observed measurement. As optimization proceeds, the predicted OCT image $\hat{I}(z, x)$ converges to the target image $I_\text{target}(z, x)$ and the recovered attenuation map $\mu(z, x)$ converges to our estimated optical attenuation coefficient at each pixel. Created in BioRender. Edelman, E. (2026) https://BioRender.com/uffg8fl}\label{fig:framework}}



\end{document}